\documentclass[11pt,letterpaper]{article}

\usepackage[
    letterpaper,
    textwidth=5in,
    top=1in,
    bottom=1in,
    hcentering
]{geometry}

\usepackage{times}
\usepackage{natbib}

\usepackage{amsmath,amsfonts,bm}

\def\eqref#1{equation~\ref{#1}}

\def\1{\bm{1}}

\DeclareMathAlphabet{\mathsfit}{\encodingdefault}{\sfdefault}{m}{sl}
\SetMathAlphabet{\mathsfit}{bold}{\encodingdefault}{\sfdefault}{bx}{n}

\usepackage{booktabs}
\usepackage{hyperref}
\usepackage{url}
\usepackage{graphicx}
\usepackage{array}
\usepackage{float}

\title{UniForge3D: Joint 3D Reconstruction and\\
Generation in a Shared Anchor-Guided Space}

\author{%
\begin{tabular}{@{}l@{}}
{\small\bfseries
Wenjue Chen$^{1,*}$ \quad Wenyue Chen$^{1,2,*}$ \quad Heliang Zheng$^{2}$}\\[-2pt]
{\small\bfseries
Rongfei Jia$^{2}$ \quad Yuan Liu$^{3,\dagger}$ \quad Meng Li$^{1,\dagger}$}\\[1pt]
{\footnotesize $^{1}$Peking University \qquad $^{2}$Math Magic}\\[-2pt]
{\footnotesize $^{3}$The Hong Kong University of Science and Technology}\\[1pt]
{\footnotesize $^{*}$Equal contribution.
\qquad $^{\dagger}$Corresponding authors.}
\end{tabular}%
}
\date{}

\makeatletter
\renewcommand{\maketitle}{%
  \par\noindent{\LARGE\bfseries\@title\par}
  \vspace{0.5em}
  \noindent\@author\par
  \vspace{0.5em}
}
\makeatother

\begin{document}

\maketitle

\begin{figure}[!h]
    \centering
    \setlength{\abovecaptionskip}{2pt}
    \includegraphics[width=\linewidth]{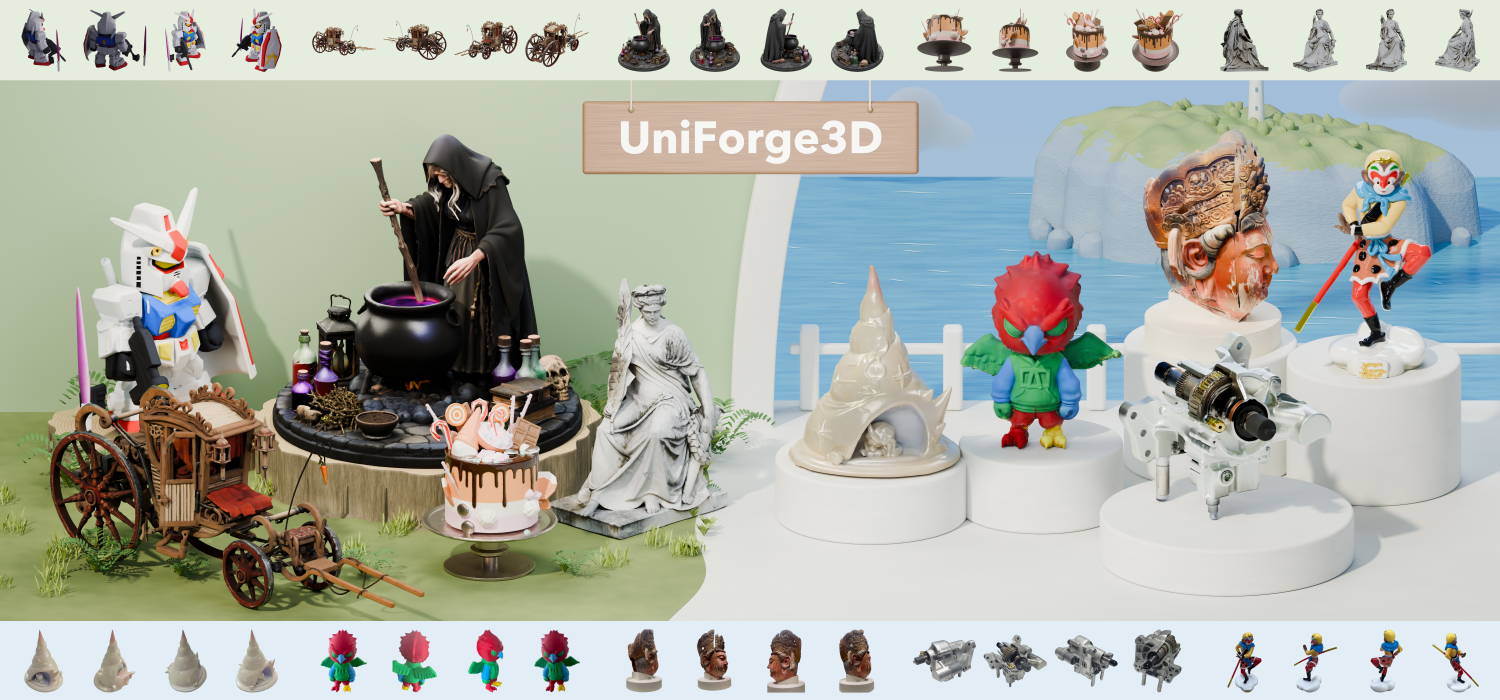}
    \caption{
        UniForge3D generates faithful 3D assets
        from sparse, unposed multi-view images.
    }
    \label{fig:teaser}
\end{figure}

\begin{abstract}
\fontsize{10}{11}\selectfont
Generating complete 3D assets from sparse, unposed images requires preserving observed geometry while completing unseen regions. Joint reconstruction and generation offers a way to address both goals: reconstruction recovers the geometry observed in the input images, while generation contributes learned shape priors. However, combining camera-relative reconstruction with canonical-space generation introduces a coordinate mismatch that makes direct geometric interaction more difficult. Moreover, conditioning each 3D location on all views without accounting for their relevance may mix useful geometric evidence with unrelated image content, hindering local detail recovery. We present \textbf{UniForge3D}, which jointly models reconstruction and generation in a shared anchor-guided space, resolving their coordinate mismatch. This space follows the anchor view's azimuth while keeping the object upright. The two branches exchange features bidirectionally to predict per-view point maps and a complete sparse voxel structure. We further condition point-map prediction on the generated structure to establish pixel-to-voxel correspondences and select the most informative input view for each voxel. Features from the selected views then guide detailed shape generation, helping preserve observed geometric details. Our method achieves state-of-the-art appearance and geometry results on Toys4K and GSO, generating complete 3D assets with improved fidelity. Project webpage: \url{https://xishuxishu.github.io/uniforge3d-project/}
\end{abstract}

\clearpage


\section{Introduction}
\label{sec:introduction}

High-quality 3D assets are essential for digital content creation, gaming, augmented and virtual reality, and embodied intelligence. In many practical scenarios, an object is observed through only a few images with unknown camera poses. From these sparse, unposed images, we aim to generate a complete 3D asset that faithfully preserves the geometry and appearance observed in every input view.

Existing multi-view 3D generation methods commonly condition native
3D generators on features extracted from multiple input images.
They incorporate multi-view evidence by assigning images to predefined view slots or adaptively fusing features across views~\citep{hunyuan3d2.0,li2026pixal3d,sun2026roar,li2026mvsam3d,zhou2026stream3d}. However, image conditioning does not strictly constrain the generated surfaces to match the observed geometry. Generative models may therefore favor plausible shapes under their learned priors over faithfully preserving input-specific structures. This motivates using multi-view reconstruction to provide more direct geometric guidance.

A straightforward approach is to reconstruct geometry from multi-view images and use the resulting implicit features or explicit geometry to condition generation~\citep{chang2026reconviagen,huang2026recgen3d,siddiqui2026shaper,hu2026axolotl3d,zhou2025gena3d,liu2026reconplusgen}. However, errors in these upstream reconstruction estimates can propagate into generation, limiting fidelity when the geometric guidance is unreliable. 

A further direction explores joint reconstruction and generation.
Mix3R~\citep{lin2026mix3r} connects pretrained reconstruction and
generation models through a Mixture-of-Transformers architecture,
allowing the two tasks to exchange information and improve each
other. However, the two branches retain different coordinate
representations, requiring an additional predicted similarity
transformation to align the reconstructed point maps with the
generated geometry.

\begin{figure}[t]
    \centering
    \includegraphics[width=\linewidth]
    {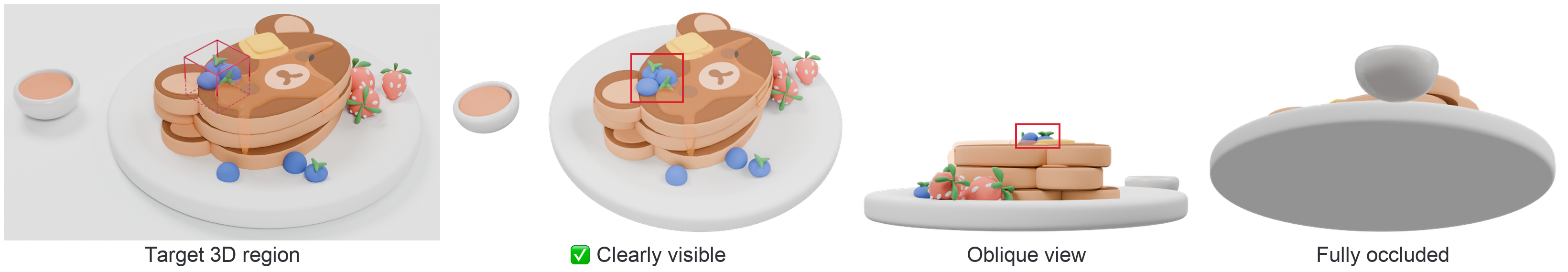}
    \vspace{-2px}
    \caption{
        \textbf{Different views provide different evidence
        for the same 3D region.}
        The 3D box marks three blueberries, which are clearly
        visible, seen at an oblique angle, or fully occluded
        across the three views, motivating region-specific
        view selection.
    }
    \label{fig:view_selection_motivation}
\end{figure}

This coordinate mismatch hinders direct geometric interaction
between reconstruction and generation, posing the first challenge. A further challenge lies in recovering local shape details from
complex multi-view observations. As illustrated in Fig.~\ref{fig:view_selection_motivation}, the same region may be fully visible with
clear details in one view, but difficult to distinguish in another
because of an oblique viewing angle or occlusion. Combining all views without accounting for these differences
can mix useful evidence with unrelated image content, making it
harder to recover details at the corresponding 3D location.

\vspace{-2px}
To address these challenges, we propose \textbf{UniForge3D},
a framework that jointly models 3D reconstruction and generation
in a shared anchor-guided space and uses structure-guided view
conditioning to recover local shape details from sparse,
unposed multi-view images.

Specifically, \textbf{UniForge3D} consists of two main components.
First, \textbf{Anchor-Guided Joint Reconstruction and Generation}
defines a shared space whose horizontal orientation follows the
anchor view's azimuth while preserving the object's upright
direction, eliminating the coordinate mismatch between the two
branches. The branches exchange features bidirectionally to predict
per-view point maps and a complete sparse voxel structure in this
space. The shared coordinate system allows image observations
to be directly mapped onto the generated structure through
the point maps. Furthermore, to establish more reliable pixel-to-voxel
correspondences, \textbf{Structure-Guided View Selection and Conditioning}
uses the generated sparse structure to guide point-map prediction.
These correspondences identify the most informative input view
for each voxel, whose features then guide local shape generation
to better preserve the observed geometric details.

Extensive experiments on the Toys4K~\citep{stojanov2021toy4k} and GSO~\citep{downs2022gso} datasets demonstrate that \textbf{UniForge3D} achieves state-of-the-art performance across all evaluated geometry and appearance metrics, generating complete 3D assets that faithfully preserve the structures and visual details observed in sparse, unposed input views.

\section{Related Work}

\subsection{Single-View 3D Generation}

Single-view 3D object generation has progressed rapidly, with existing approaches broadly categorized into methods based on 2D generative priors and native 3D generative methods. Within the former category, one paradigm optimizes 3D representations under supervision distilled from pretrained 2D diffusion models~\citep{poole2022dreamfusion,lin2023magic3d,wang2023prolificdreamer,melas2023realfusion,tang2023make,qian2024magic123,tang2024dreamgaussian,liu2023zero}. The other paradigm employs image or video diffusion models to synthesize multiple views of an object and aggregates the resulting multi-view information to recover 3D representations~\citep{shi2023zero123++,liu2024syncdreamer,long2024wonder3d,xu2024instantmesh,tang2024lgm,xu2024grm,liu2023one2345,li2024era3d,zuo2024videomv,wu2024unique3d}. 

Recent research has increasingly shifted toward native 3D generation, where generative models are trained directly on large-scale 3D data. Driven by advances in 3D latent representations, current approaches primarily follow two paradigms: Vector Set (VecSet)~\citep{zhang20233dshape2vecset,li2024craftsman3d,zhang2024clay,li2025triposg,hunyuan3d2025hunyuan3d21,li2025step1x} and Sparse Voxel~\citep{ren2024xcube,xiang2025trellis,wu2026direct3d,he2025sparseflex,chen2026synchuman,li2026sparc3d,xiang2026trelli2,chen2026know3d,ye2025hi3dgen}. VecSet-based methods encode 3D shapes as collections of latent vectors, providing compact representations and effective global modeling, whereas Sparse-Voxel-based methods retain explicit spatial organization and better support local geometric details. 

\subsection{Multi-View 3D Reconstruction and Generation}

Classical multi-view reconstruction combines Structure-from-Motion and multi-view stereo~\citep{schonberger2016structure,furukawa2010mvs}, while learning-based methods improve depth estimation through learned matching~\citep{yao2018mvsnet,gu2020cascade,wang2021patchmatchnet}. Neural rendering and feed-forward Gaussian prediction further support reconstruction and novel-view synthesis~\citep{mildenhall2021nerf,niemeyer2022regnerf,charatan2024pixelsplat,zhang2025transplat}. Recent feed-forward models directly estimate point maps and other geometric quantities from unposed images~\citep{wang2024dust3r,leroy2024mast3r,yang2025fast3r,wang2025cut3r,wang2025vggt,wang2026pi3,keetha2026mapanything,lin2025da3,wang2026vggtOmega}. However, their reconstructions primarily cover observed surfaces and may remain incomplete in unseen regions.

Multi-view 3D generation follows two main approaches. Training-free methods extend pretrained single-view generators by adaptively selecting or fusing information across views~\citep{li2026mvsam3d,zhou2026stream3d,qu2026umi3d}. Training-based methods learn to incorporate multiple views through feature conditioning or adaptive view routing~\citep{hunyuan3d2.0,li2026pixal3d,sun2026roar}. These strategies improve the use of multi-view evidence, but image conditioning alone does not ensure that generated surfaces faithfully match the observed geometry.

To combine reconstruction with generative priors, one line of work uses estimated reconstruction features or geometry as generation conditions~\citep{chang2026reconviagen,huang2026recgen3d,siddiqui2026shaper,hu2026axolotl3d,zhou2025gena3d,liu2026reconplusgen}. For example, ReconViaGen uses reconstruction features, while RecGen3D additionally incorporates canonicalized point clouds. Such guidance provides geometric and cross-view information, but errors in the reconstruction estimates may propagate into generation. Another direction jointly models reconstruction and generation. Mix3R~\citep{lin2026mix3r} enables bidirectional interaction between the two branches, but retains camera-relative reconstruction and canonical-space generation. It therefore requires learning cross-space alignment alongside both tasks, complicating direct geometric interaction between the two branches. Our method instead places reconstruction and sparse-structure generation in a shared anchor-guided space, giving both branches a consistent geometric reference. We further use structure-guided reconstruction to identify informative input views for detailed shape generation.

\section{Method}
\label{sec:method}

\begin{figure}[t]
    \centering
    \includegraphics[width=\linewidth]{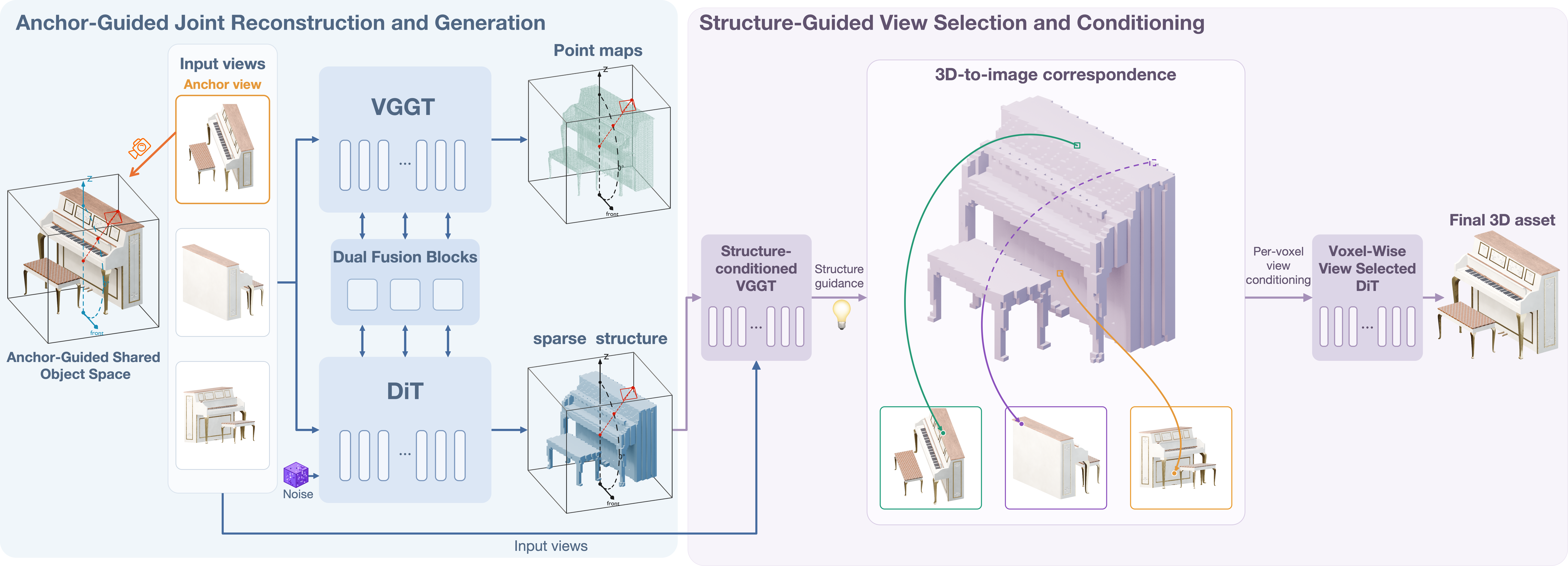}
    \caption{
        \textbf{Overview of UniForge3D.}
        \textbf{Left:} The reconstruction and generation branches
        interact bidirectionally to predict multi-view point maps
        and a complete sparse structure in a shared anchor-guided space.
        \textbf{Right:} Structure-conditioned point maps establish
        voxel-to-pixel correspondences, enabling the model to select
        informative views and use their image features to guide
        detailed shape generation.
    }
    \label{fig_pipeline}
\end{figure}
\paragraph{Overview.}
Given a sparse set of unposed images
$\mathcal{I}=\{I_i\}_{i=1}^{N}$ of an object, we aim to generate
a complete 3D asset that faithfully preserves the geometry
observed in every input view. We designate $I_1$ as the anchor
view and jointly predict multi-view point maps and a complete
coarse voxel structure in an anchor-guided shared object space
(Section~\ref{sec:anchor_shared_space}).
The shared coordinate system further allows reconstructed
3D points to be directly associated with generated voxels.
We therefore use structure-guided reconstruction to establish
explicit pixel-to-voxel correspondences and select the most
informative view for each voxel.
The selected-view image features then guide detailed shape
generation (Section~\ref{sec:structure_correspondence}).
The overall pipeline is illustrated in Fig.~\ref{fig_pipeline}.

\subsection{Anchor-Guided Joint Reconstruction and Generation}
\label{sec:anchor_shared_space}

To achieve high-fidelity multi-view 3D object generation, we seek to make reconstruction and generation mutually beneficial. Directly conditioning generation on reconstructed geometry or features can propagate reconstruction errors. Joint training enables mutual assistance, but their different coordinate systems complicate geometric interaction: feed-forward reconstruction~\citep{wang2025vggt} predicts geometry in the first-camera space, whereas native 3D generation~\citep{xiang2026trelli2} operates in a canonical object space. We therefore introduce a shared anchor-guided space that follows the anchor view's azimuth while preserving the canonical vertical direction. Both branches operate in this shared space and exchange features bidirectionally, so that observed geometry guides structure generation and complete-shape context improves reconstruction.

\subsubsection{Anchor-Guided Shared Object Space}

We define a shared object space whose front direction follows the azimuth of the anchor view $I_1$, while its up direction remains aligned with the canonical vertical axis. For any object point $\mathbf{p}^{\mathrm{can}}$ in the centered and scale-normalized canonical space, its shared-space coordinate is
\begin{equation}
    \mathbf{p}^{\star}
    =
    R_a\mathbf{p}^{\mathrm{can}},
    \label{eq:anchor_shared_transform}
\end{equation}
where $R_a$ rotates about the canonical vertical axis so that the anchor's
viewing azimuth defines the front direction in the shared space.

We construct the targets of both branches from the transformed geometry. The visible surface points form the per-view point-map targets for reconstruction, while the complete geometry is voxelized to form the sparse-structure target for generation. Consequently, the two branches describe geometry in the same object space and can interact without additional alignment.

\subsubsection{Bidirectional Joint Modeling}

The two branches contain complementary geometric information. The reconstruction branch describes surfaces observed in the input views, while the generation branch models the complete sparse structure. We connect them through bidirectional interactions at selected layers. We denote the generation tokens by $\mathbf{G}^{l}_{t}$ and the reconstruction tokens of view $v$ by $\mathbf{R}^{l}_{v}$.

\paragraph{Reconstruction-to-generation.}
For each view, we spatially align the reconstruction tokens with the DINOv3~\citep{simeoni2025dinov3} patch features. The reconstruction features predict a residual that is added to the corresponding DINOv3 features, producing a view-specific condition $\mathbf{C}^{l}_{v}$. We then concatenate the conditions from all views along the sequence dimension:
\begin{equation}
\mathbf{C}^{l}_{\mathrm{mv}}
=
\operatorname{Concat}_{v}
\left[
\mathbf{D}_{v}
+
\phi^{l}\!\left(\mathbf{R}^{l}_{v}\right)
\right].
\end{equation}
The generation tokens attend to $\mathbf{C}^{l}_{\mathrm{mv}}$ through the cross-attention layers of the generation branch. This allows multi-view geometric and semantic evidence to guide sparse-structure generation.

\paragraph{Generation-to-reconstruction.}
After the corresponding generation blocks, the reconstruction tokens from all views are concatenated along the sequence dimension and attend to the updated generation tokens:
\begin{equation}
\widetilde{\mathbf{R}}^{l}
=
\mathbf{R}^{l}
+
g^{l}(t)\odot
\mathcal{A}^{l}_{g\rightarrow r}
\left(
Q=\mathbf{R}^{l},
K=V=\widetilde{\mathbf{G}}^{l}_{t}
\right),
\end{equation}
where $g^{l}(t)$ gates the injected features according to the noise level. This interaction injects global shape context into the reconstruction
branch, helping it estimate the 3D positions of observed pixels.

We jointly optimize the two branches using
\begin{equation}
\mathcal{L}
=
\mathcal{L}_{\mathrm{flow}}
+
\mathcal{L}_{\mathrm{point}}.
\end{equation}
Since both branches are supervised using geometry expressed in the same anchor-guided space, the two branches can exchange geometric information without an additional rigid alignment.

\subsection{Structure-Guided View Selection and Conditioning}
\label{sec:structure_correspondence}

Given a generated sparse voxel structure $\mathcal{V}$ and the input images $\mathcal{I}$, this stage generates a
fine-grained shape. A surface region may be clearly visible in
one view but occluded or difficult to distinguish in another.
To identify the observations relevant to each voxel, we perform
structure-guided reconstruction in the shared anchor-guided space.
The predicted point maps provide per-pixel 3D coordinates in the
same coordinate system as the generated voxels, enabling explicit
pixel-to-voxel correspondences. These correspondences allow us
to select the most informative view for each voxel based on valid
pixel support and use its image features to guide detailed shape
generation.



\subsubsection{Structure-Guided Reconstruction}

We use the generated sparse structure $\mathcal{V}$ as an additional
condition for reconstruction, allowing the network to use both image
observations and the current generated geometry when estimating
per-pixel 3D positions. We encode $\mathcal{V}$ into a sequence of
tokens $\mathbf{S}$, which the reconstruction network attends to
through cross-attention at selected layers. The network predicts point maps
$\hat{\mathbf{P}}_v^{\mathcal{V}} \in \mathbb{R}^{H \times W \times 3}$
and confidence maps
$\hat{\rho}_v^{\mathcal{V}} \in \mathbb{R}^{H \times W}$,
along with reconstruction features for subsequent image conditioning.
The predicted points share the coordinate system of $\mathcal{V}$,
enabling direct association between image pixels and generated voxels.

\subsubsection{Voxel-Wise View Selection and Conditioning}

We use the predicted point maps to establish pixel-to-voxel
correspondences directly in the shared anchor-guided space.
After confidence filtering, we quantize the predicted points
onto the voxel grid. For each voxel centered at $x_j$ and each
view $v$, we identify the retained pixels mapped to that voxel.
We define the support score $s_v(x_j)$ as the number of pixels
in the largest connected component on the image. A larger score suggests that the view provides
more visual information about the region represented by the
voxel. For each voxel, we select
\begin{equation}
    v_j^{\star} = \arg\max_v s_v(x_j).
    \label{eq:voxel_view_selection}
\end{equation}

Voxels without direct support inherit a view from a directly supported voxel in their 26-neighborhood; if none exists, they use the anchor view. 

For each view, an adapter aligns the Structure-Guided Reconstruction features with the DINOv3~\citep{simeoni2025dinov3} patch tokens. The aligned features are added to the corresponding DINOv3 tokens to form the view-specific feature sequence. The shape denoiser interleaves two types of cross-attention: \textbf{global multi-view attention}, in which every voxel token attends to the features of all views concatenated along the sequence dimension, and \textbf{voxel-specific selected-view attention}, in which each voxel token attends only to its selected view $v_j^\star$. The former provides global context for shape completion, while the latter supplies reliable local evidence.

\section{Experiments}
\label{sec:Experiments}

\subsection{Experimental Setup}
\paragraph{Data Preparation.}
Our model is trained on a subset of the TRELLIS-500K
dataset~\citep{xiang2025trellis} containing 210K objects.
For each object, we sample two random azimuthal orientations
and render six candidate anchor views for each orientation.
We additionally render 48 views from randomly sampled viewpoints
to provide additional input views for multi-view training.
We randomly select 500 objects from Toys4K~\citep{stojanov2021toy4k}
and 500 from Google Scanned Objects (GSO)~\citep{downs2022gso},
for a total of 1,000 evaluation objects.
All test objects are held out from training. For each test object,
we render four images from randomly selected viewpoints. All methods
receive the same four unposed RGB images.

\paragraph{Evaluation Metrics.}
We evaluate appearance quality using PSNR, SSIM~\citep{wang2004ssim}, and LPIPS~\citep{zhang2018lpips}. Chamfer
Distance (CD)~\citep{fan2017cd} is computed as the sum of the bidirectional mean
nearest-neighbor distances between points sampled from the reconstructed
and ground-truth surfaces. We compute F-score using a distance threshold
of $0.01$. All appearance and geometry metrics are computed after aligning
each reconstructed mesh with its ground-truth mesh using a GICP-based
Sim(3)~\citep{segal2009generalized} transformation.
For the ablation on anchor-guided joint modeling, each predicted sparse
structure is aligned with the ground truth before evaluating the $64^3$
binary surface-occupancy grids using IoU and CD. For point-map
reconstruction, predictions from all input views are jointly aligned with
the visible ground-truth position maps using a shared object-level Umeyama
Sim(3) transformation~\citep{umeyama1991least}. Accuracy measures the mean distance from predicted
to ground-truth points, Completeness measures the mean distance in the
reverse direction, and Overall is their arithmetic mean. For the
view-conditioning ablation, PSNR, SSIM, and LPIPS are computed on rendered
normal maps.

\paragraph{Training Details.}
All models are trained on 32 NVIDIA A800 GPUs.
For Anchor-Guided Joint Reconstruction and Generation,
we initialize the generation and reconstruction branches
from pretrained TRELLIS.2~\citep{xiang2026trelli2} and
VGGT~\citep{wang2025vggt}, respectively, and fine-tune them
using LoRA~\citep{hu2021lora} with ranks 64 and 32,
while fully training the new interaction modules.
We train for 60K iterations with a global batch size of 128
and a learning rate of $5\times10^{-5}$.
For Structure-Guided Reconstruction, we fully fine-tune
pretrained VGGT for 40K iterations, then freeze it during
subsequent generator training.
For coarse-to-fine shape generation, we fine-tune pretrained
TRELLIS.2 denoisers using rank-128 LoRA and fully train the
new modules. Each model is trained for 60K iterations with
a global batch size of 64 and a learning rate of $5\times10^{-5}$.
The texture generator uses the same training settings as
the detailed shape generator.
\begin{figure*}[p]
    \centering
    \begin{minipage}{0.995\textwidth}
        \centering
        \includegraphics[width=\linewidth]{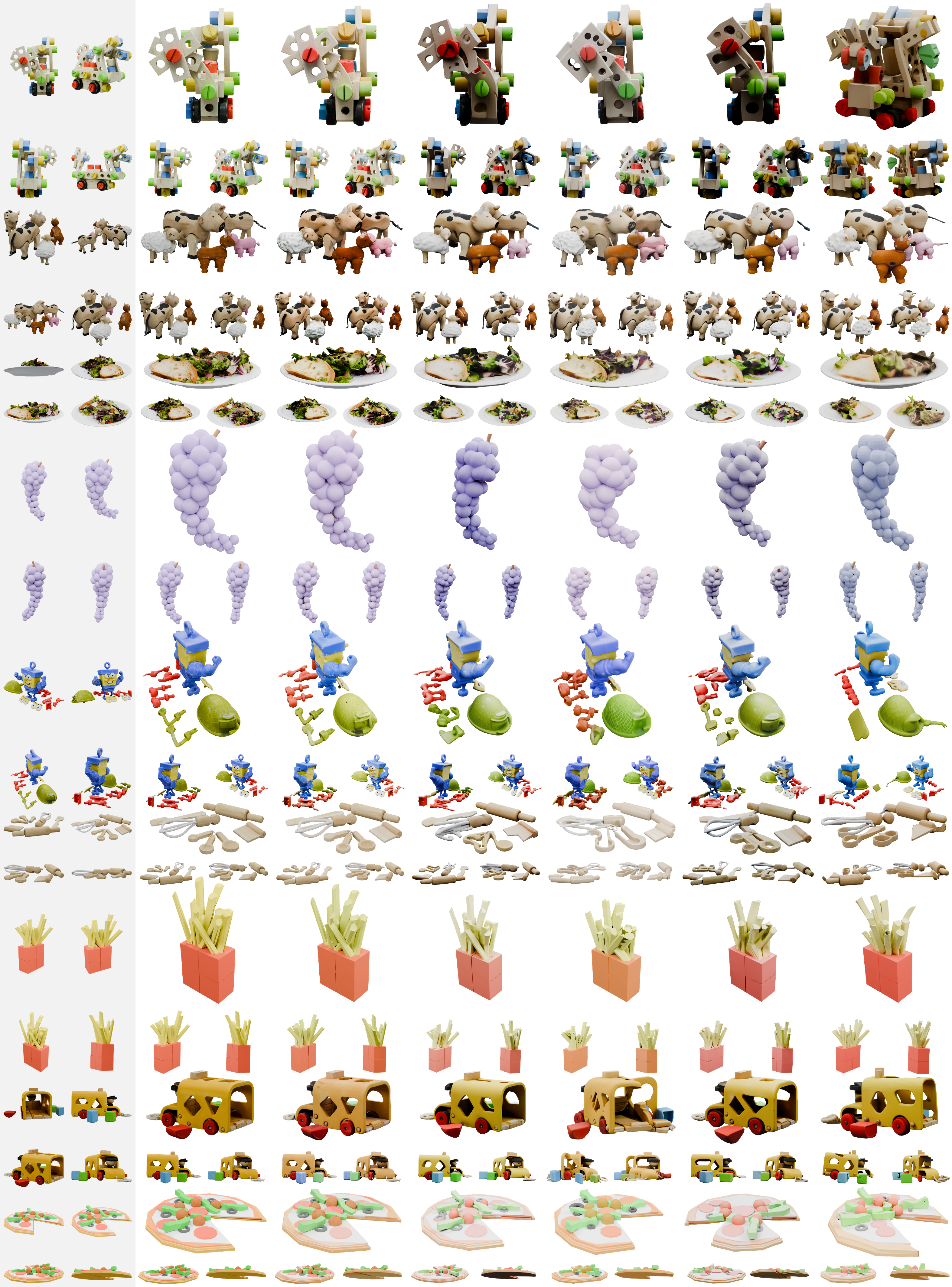}

        \noindent\makebox[\linewidth][l]{%
            \parbox{0.14285\linewidth}{\centering\scriptsize Input}%
            \parbox{0.14285\linewidth}{\centering\scriptsize GT}%
            \parbox{0.14285\linewidth}{\centering\scriptsize Ours}%
            \parbox{0.14285\linewidth}{\centering\scriptsize Mix3R}%
            \parbox{0.14285\linewidth}{\centering\scriptsize ReconViaGen\\v0.5}%
            \parbox{0.14285\linewidth}{\centering\scriptsize ReconViaGen\\v0.2}%
            \parbox{0.14285\linewidth}{\centering\scriptsize MV-SAM3D}%
        }
    \end{minipage}
    \caption{
        Qualitative comparisons on Toys4K and GSO.
        Each row shows four input views, ground-truth renderings, and
        results from different methods.
        To enable direct comparison, each generated mesh is geometrically
        aligned with the ground-truth asset and rendered using the
        corresponding input-view cameras.
        Our method better preserves the observed appearance and local
        structures while producing complete and consistent geometry.
    }
    \label{fig:qualitative_synthetic}
\end{figure*}

\subsection{Comparison with State-of-the-Art Methods}
\paragraph{Quantitative Results.}
We compare our method with Hunyuan3D-2MV~\citep{hunyuan3d2.0},
Mix3R~\citep{lin2026mix3r}, MV-SAM3D~\citep{li2026mvsam3d},
ReconViaGen~\citep{chang2026reconviagen}, and
Stream3D + TRELLIS.2~\citep{zhou2026stream3d}. As shown in Table~\ref{tab:main_comparison}, our method achieves the best
performance on both Toys4K~\citep{stojanov2021toy4k} and
GSO~\citep{downs2022gso} across all appearance and geometry metrics.
The improvements in CD and F-score demonstrate more accurate geometry, while
the gains in PSNR, SSIM, and LPIPS indicate better appearance preservation
under novel viewpoints.

\begin{figure*}[t]
    \centering
    \includegraphics[width=\textwidth]{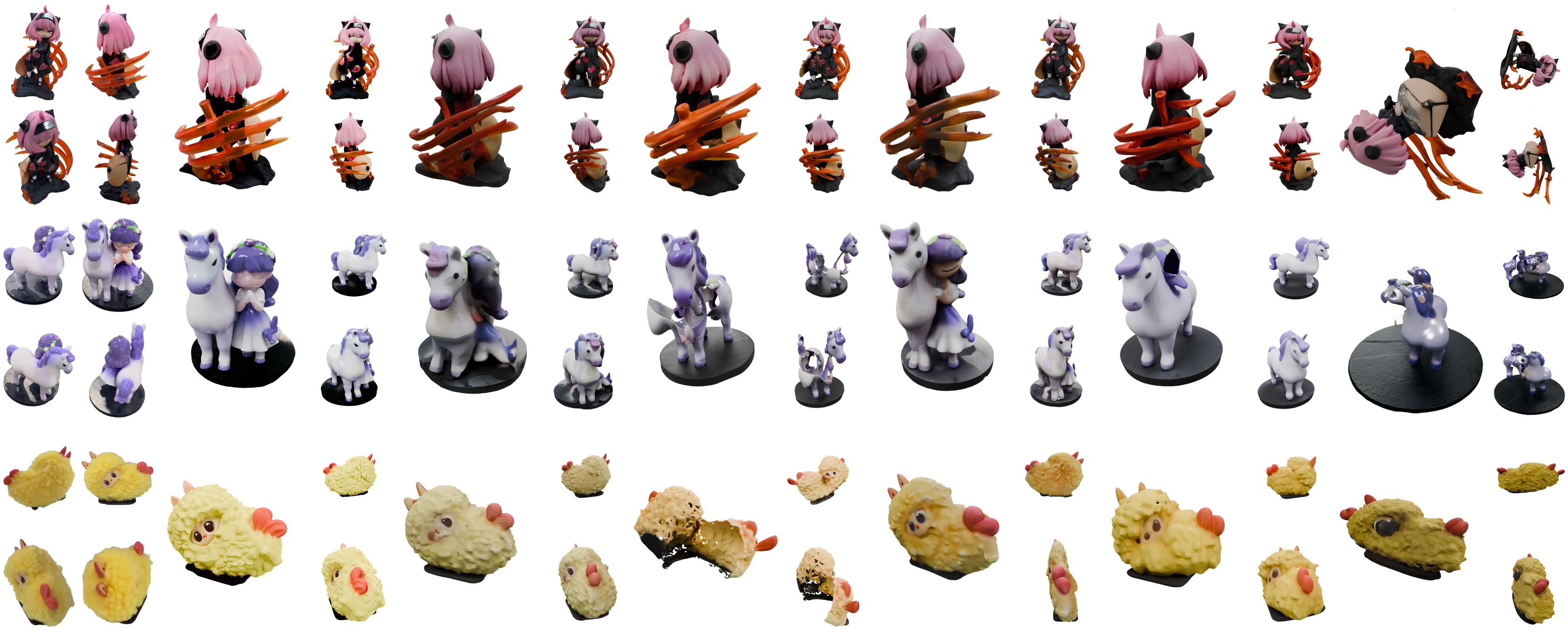}

    \noindent\makebox[\textwidth][l]{%
        \parbox{0.10\textwidth}{\centering\scriptsize Input}%
        \parbox{0.15\textwidth}{\centering\scriptsize Ours}%
        \parbox{0.15\textwidth}{\centering\scriptsize Mix3R}%
        \parbox{0.15\textwidth}{\centering\scriptsize ReconViaGen\\v0.5}%
        \parbox{0.15\textwidth}{\centering\scriptsize ReconViaGen\\v0.2}%
        \parbox{0.15\textwidth}{\centering\scriptsize MV-SAM3D}%
        \parbox{0.15\textwidth}{\centering\scriptsize Stream3D}%
    }
    \caption{
        Qualitative comparisons on real-world sparse, unposed images. All results are aligned and rendered from common viewpoints for
        visual comparison.
    }
    \label{fig:qualitative_realworld}
\end{figure*}

\paragraph{Qualitative Results.}
Fig.~\ref{fig:qualitative_synthetic}
and Fig.~\ref{fig:qualitative_realworld} compare results on synthetic
benchmarks and real-world images, respectively.
Our method better preserves observed geometry and appearance,
while competing methods often omit parts or distort local
structures. These results demonstrate its fidelity to the input observations.

\begin{table*}[t]
\centering
\caption{Quantitative comparison on the Toys4K~\citep{stojanov2021toy4k} and GSO~\citep{downs2022gso} datasets.
Best results are shown in bold and second-best results are underlined. PSNR, SSIM, and LPIPS are computed on RGB renderings.}
\label{tab:main_comparison}

\setlength{\tabcolsep}{4pt}
\renewcommand{\arraystretch}{1.08}

\resizebox{\textwidth}{!}{
\begin{tabular}{l|ccccc|ccccc}
    \hline
    \textbf{Dataset}
    & \multicolumn{5}{c|}{\textbf{Toys4K~\citep{stojanov2021toy4k}}}
    & \multicolumn{5}{c}{\textbf{GSO~\citep{downs2022gso}}} \\
    \hline

    \textbf{Metric}
    & PSNR$\uparrow$
    & SSIM$\uparrow$
    & LPIPS$\downarrow$
    & CD ($\times 10^{-2}$)$\downarrow$
    & F-score$\uparrow$
    & PSNR$\uparrow$
    & SSIM$\uparrow$
    & LPIPS$\downarrow$
    & CD ($\times 10^{-2}$)$\downarrow$
    & F-score$\uparrow$ \\
    \hline

    Stream3D + TRELLIS.2~\citep{zhou2026stream3d}
    & 20.955 & 0.9351 & 0.0789 & 6.3608 & 0.5003
    & 19.576 & 0.9119 & 0.0990 & 6.2441 & 0.4700 \\

    Hunyuan3D-2MV~\citep{hunyuan3d2.0}
    & 22.726 & 0.9470 & 0.0576 & 2.9175 & 0.6723
    & 20.526 & 0.9206 & 0.0780 & 3.2713 & 0.6139 \\

    MV-SAM3D~\citep{li2026mvsam3d}
    & 23.071 & 0.9469 & 0.0590 & 2.7147 & 0.6830
    & 20.653 & 0.9222 & 0.0791 & 2.7447 & 0.6517 \\

    ReconViaGen v0.2~\citep{chang2026reconviagen}
    & 22.713 & 0.9451 & 0.0548 & 2.6155 & 0.6786
    & 20.498 & 0.9196 & 0.0738 & 2.5905 & 0.6697 \\

    ReconViaGen v0.5~\citep{chang2026reconviagen}
    & \underline{25.020}
    & \underline{0.9553}
    & 0.0519
    & 2.7007
    & 0.6840
    & \underline{22.774}
    & \underline{0.9313}
    & 0.0731
    & 2.7002
    & 0.6489 \\

    Mix3R~\citep{lin2026mix3r}
    & 23.907
    & 0.9498
    & \underline{0.0506}
    & \underline{2.3445}
    & \underline{0.7343}
    & 21.040
    & 0.9239
    & \underline{0.0688}
    & \underline{2.3202}
    & \underline{0.7174} \\

    \textbf{Ours}
    & \textbf{27.428}
    & \textbf{0.9616}
    & \textbf{0.0401}
    & \textbf{1.4196}
    & \textbf{0.8899}
    & \textbf{24.826}
    & \textbf{0.9363}
    & \textbf{0.0591}
    & \textbf{1.4047}
    & \textbf{0.8897} \\

    \hline
\end{tabular}
}
\end{table*}

\subsection{Ablation Study}

\paragraph{Ablation on Anchor-Guided Joint Modeling.}Table~\ref{tab:sparse_structure_gso} compares different conditioning
schemes and coordinate spaces for sparse-structure generation. \textit{DINO Conditioning} concatenates per-view
DINOv3~\citep{simeoni2025dinov3} features along the sequence
dimension, with separate embeddings for anchor and non-anchor
views. \textit{DINO--VGGT Conditioning} additionally uses
VGGT~\citep{wang2025vggt}, fine-tuned for
point-map prediction in our anchor-guided shared space and
then frozen. An MLP adapter with a zero-initialized output
projection fuses spatially aligned features from four VGGT
layers into a residual, which is added to DINOv3 features
before concatenating views. Its consistent gains over DINO
Conditioning show that reconstruction features complement
DINOv3 with geometric cues and cross-view relationships. Our full model, \textit{Anchor-Guided Joint Training}, further
jointly optimizes the reconstruction and generation branches
through bidirectional feature interaction. It achieves the best
results across all metrics, supporting the benefit of joint
optimization with bidirectional feature interaction. 

We further compare coordinate-space choices for joint training.
In \textit{Canonical-Space Joint Training}, both branches operate
in the canonical object space. In \textit{Cross-Space Joint Training},
generation operates in the canonical object space while
reconstruction remains in VGGT's first-camera space.  Both variants perform worse than our anchor-guided formulation,
highlighting the benefit of a shared coordinate space whose
orientation is tied to the input views.

\begin{table*}[!h]
\centering
\caption{Ablation of conditioning schemes and coordinate spaces for sparse-structure generation on GSO~\citep{downs2022gso}.}
\label{tab:sparse_structure_gso}
\resizebox{\textwidth}{!}{
\begin{tabular}{lllcc}
\toprule
Method
& VGGT Space
& DiT Space
& IoU$\uparrow$
& CD ($\times 10^{-2}$)$\downarrow$ \\
\midrule

DINO Conditioning
& N/A
& Anchor-Guided Space
& 0.4261
& 2.0019 \\

DINO--VGGT Conditioning
& Anchor-Guided Space
& Anchor-Guided Space
& \underline{0.5533}
& \underline{1.1710} \\

Canonical-Space Joint Training
& Canonical Space
& Canonical Space
& 0.4601
& 1.5825 \\

Cross-Space Joint Training
& First-Camera Space
& Canonical Space
& 0.4563
& 1.6291 \\

Anchor-Guided Joint Training (Ours)
& Anchor-Guided Space
& Anchor-Guided Space
& \textbf{0.5827}
& \textbf{1.0563} \\

\bottomrule
\end{tabular}
}
\end{table*}
\begin{table}[!h]

    \centering
\caption{
Point-map reconstruction results on the GSO dataset~\citep{downs2022gso}.
Distances are reported in normalized object coordinates
($\times 10^{-3}$).
}
    \label{tab:gso_pointmap}
    \small
    \setlength{\tabcolsep}{8pt}
    \begin{tabular}{lccc}
        \toprule
        Method
        & Acc.$\downarrow$
        & Comp.$\downarrow$
        & Overall$\downarrow$ \\
        \midrule
        Image-Only VGGT
        & 5.025
        & 4.667
        & 4.846 \\

        Jointly Trained VGGT
        & \underline{4.053}
        & \underline{4.250}
        & \underline{4.152} \\

        Structure-Conditioned VGGT
        & \textbf{3.539}
        & \textbf{3.086}
        & \textbf{3.313} \\
        \bottomrule
    \end{tabular}
\end{table}

\vspace{-8px}

\begin{table}[!h]

\centering
\caption{
Ablation of view-conditioning strategies for detailed shape generation
on GSO~\citep{downs2022gso}. Both variants take the same predicted $64^3$ sparse structure
as input. PSNR, SSIM, and LPIPS are computed on rendered normal maps.
}
\label{tab:view_conditioning_ablation_gso}

\setlength{\tabcolsep}{5pt}
\renewcommand{\arraystretch}{1.08}

\resizebox{\columnwidth}{!}{
\begin{tabular}{lccccc}
\hline
\textbf{Method}
& PSNR$\uparrow$
& SSIM$\uparrow$
& LPIPS$\downarrow$
& CD ($\times 10^{-2}$)$\downarrow$
& F-score$\uparrow$ \\
\hline

Full Multi-View Conditioning
& 26.892
& 0.9649
& 0.0401
& 1.4739
& 0.8758 \\

Voxel-Wise View Selection and Conditioning (Ours)
& \textbf{27.207}
& \textbf{0.9657}
& \textbf{0.0388}
& \textbf{1.4047}
& \textbf{0.8897} \\

\hline
\end{tabular}
}
\end{table}

\paragraph{Ablation on Reconstruction Modeling.}
We ablate the reconstruction design using three VGGT variants.
\textit{Image-Only VGGT} is independently fine-tuned end-to-end using
only multi-view images.
\textit{Jointly Trained VGGT} is jointly optimized with the
sparse-structure generator through bidirectional feature interaction
in the first stage.
\textit{Structure-Conditioned VGGT} further introduces the generated
sparse structure as an explicit condition for point-map prediction in
the second stage. As shown in Table~\ref{tab:gso_pointmap},
\textit{Jointly Trained VGGT} outperforms
\textit{Image-Only VGGT}, supporting the benefit of interaction with
the generation branch for reconstruction.
\textit{Structure-Conditioned VGGT} produces the most accurate and
complete point maps, demonstrating the benefit of explicit
sparse-structure conditioning for reconstruction.

\paragraph{Ablation on Voxel-Wise View Selection and Conditioning.}
Using the same predicted sparse structure, we compare two conditioning strategies for detailed shape generation. In \textbf{Full Multi-View Conditioning}, the adapter-aligned reconstruction features are added to the corresponding DINOv3 patch tokens for each view. The resulting feature sequences from all views are concatenated along the sequence dimension and used to condition every denoiser layer. Our \textbf{Voxel-Wise View Selection and Conditioning} alternates full multi-view layers with voxel-specific selected-view layers. As shown in Table~\ref{tab:view_conditioning_ablation_gso}, our design improves geometric reconstruction performance. These results show our voxel-wise view selection and conditioning strategy improves the fidelity of generated shapes.

\section{Limitation and Conclusion}
We presented UniForge3D for joint 3D reconstruction and generation from sparse, unposed images. The two branches exchange features in a shared anchor-guided space to predict per-view point maps and a complete sparse structure. The generated structure then guides reconstruction to establish pixel-to-voxel correspondences and select informative views for detailed shape generation. Quantitative results on Toys4K and GSO (Table~\ref{tab:main_comparison}), qualitative comparisons in Figs.~\ref{fig:qualitative_synthetic} and~\ref{fig:qualitative_realworld}, and the supplementary video demonstrate that UniForge3D generates complete 3D assets with faithful geometry and appearance and generalizes well to real-world images.

\textbf{Limitation.}
Our model is trained on synthetic renderings, and severe occlusion, strong reflections, and extreme lighting remain challenging. Future work will explore training with simulations of these challenging conditions to improve real-world robustness.


\bibliography{uniforge3d}
\bibliographystyle{uniforge3d}

\appendix
\section{Appendix}

\subsection{More Details}
\label{app:details}

\subsubsection{Training Details}

\paragraph{Anchor-Guided Joint Reconstruction and Generation.} We train the joint model on 32 NVIDIA A800 GPUs with a global batch size of 128. The generation and reconstruction branches are built upon pretrained TRELLIS.2~\citep{xiang2026trelli2} and VGGT~\citep{wang2025vggt} models, respectively, and are adapted using LoRA~\citep{hu2021lora} with ranks 64 and 32. We fully train the newly introduced interaction modules, with zero-initialized condition-residual outputs and time-gating projections, and keep DINOv3~\citep{simeoni2025dinov3} frozen. We use a peak learning rate of \(5\times10^{-5}\). To enable classifier-free guidance, we randomly drop the image condition with probability 0.1; the same strategy is used in subsequent generative stages.

The training objective combines latent flow matching with point-map supervision:

\begin{equation}
\label{eq:details-1}
\begin{aligned}
\mathcal L_{\mathrm{joint}}
&=\mathcal L_{\mathrm{flow}}^{\mathrm{structure}}+\mathcal L_{\mathrm{point}},\\
\mathcal L_{\mathrm{point}}
&=\mathcal L_{\mathrm{reg}}+\mathcal L_{\mathrm{conf}}
+\mathcal L_{\mathrm{normal}}.
\end{aligned}
\end{equation}

Here, \(\mathcal L_{\mathrm{flow}}\) is the mean squared error between the predicted and target latent velocities. For a valid foreground pixel, let \(e_i=\|\widehat P_i-P_i\|_2\) denote the point error and \(\widehat\rho_i\) the predicted confidence. The regression and confidence terms are

\begin{equation}
\label{eq:details-2}
\mathcal L_{\mathrm{reg}}=\mathcal R(\{e_i\}),
\qquad
\mathcal L_{\mathrm{conf}}
=\mathcal R\big(\{\widehat\rho_i e_i-0.2\log\widehat\rho_i\}\big),
\end{equation}

where \(\mathcal R\) denotes robust averaging with a 98\% quantile filter. \(\mathcal L_{\mathrm{normal}}\) measures the mean cosine error between point-map normals at three scales.

\paragraph{Structure-Guided Reconstruction.} We train this network on 32 NVIDIA A800 GPUs, using dynamic batches of up to 24 images per GPU. We initialize the model from pretrained VGGT and fully fine-tune it with a peak learning rate of \(5\times10^{-5}\). The training objective is

\begin{equation}
\label{eq:details-3}
\mathcal L_{\mathrm{rec}}
=\mathcal L_{\mathrm{reg}}+\mathcal L_{\mathrm{conf}}
+\mathcal L_{\mathrm{normal}}.
\end{equation}

After training, this network is frozen to provide point maps and reconstruction features for shape and texture generation.

\paragraph{Structure-Guided Shape Generation.} We train the 512 and 1024 shape generators separately on 32 NVIDIA A800 GPUs with a global batch size of 64. Both models are initialized from pretrained TRELLIS.2~\citep{xiang2026trelli2} shape generators, with rank-128 LoRA fine-tuning of the denoisers and full optimization of the feature adapters, while DINOv3 and the Structure-Guided Reconstruction network remain frozen. Both stages use a peak learning rate of $5\times10^{-5}$. Both stages use latent flow matching on precomputed shape latents, with \(\mathcal L_{\mathrm{shape}}=\mathcal L_{\mathrm{flow}}^{\mathrm{shape}}\).

\paragraph{Shape-Conditioned Texture Generation.} We fine-tune the texture generator on 32 NVIDIA A800 GPUs with a global batch size of 64 and a peak learning rate of \(5\times10^{-5}\). We initialize it from a previously trained TRELLIS.2-based texture model and optimize rank-128 denoiser LoRA parameters together with the full feature adapter. The reconstruction network, DINOv3, and latent autoencoders remain frozen. During training, we keep the shape latent fixed and apply flow matching to the texture latent.

All stages use AdamW with cosine learning-rate decay. Each training sample contains four views. In joint training, point-map supervision is restricted to samples retaining the image condition.

\subsubsection{Detailed Network Structure}

\paragraph{Anchor-Guided Joint Reconstruction and Generation.} The joint model comprises a VGGT reconstruction branch and a DiT generation branch. They exchange features bidirectionally in the shared anchor-guided space to jointly predict per-view point maps and a complete sparse structure.

\noindent\textbf{(1) Reconstruction-to-generation.} For each view \(v\), we concatenate the frame-wise and cross-view VGGT features into \(F_v\), map them through an adapter \(\phi\), and add the result to the spatially corresponding DINOv3 image features \(D_v\). We then concatenate all views along the sequence dimension:

\begin{equation}
\label{eq:details-4}
C_v=D_v+\phi(F_v),
\qquad
C_{\mathrm{mv}}=\operatorname{Concat}_v[C_v].
\end{equation}

DiT uses generation features as queries and \(C_{\mathrm{mv}}\) as keys and values in cross-attention, allowing generation to use the observed geometry provided by the reconstruction branch.

\noindent\textbf{(2) Generation-to-reconstruction.} After updating the generation features, we concatenate the global VGGT features across views into \(R\) and use them as queries, with generation features \(G\) as keys and values:

\begin{equation}
\label{eq:details-5}
R'=R+\operatorname{CA}_t(R,G,G).
\end{equation}

Here, \(\operatorname{CA}_t\) denotes cross-attention with time-conditioned modulation and gating, allowing the reconstruction branch to receive complete-shape information according to the current denoising state. The updated \(R'\) continues through reconstruction. After repeated bidirectional interactions, the two branches output point maps and a sparse structure through the point head and structure decoder, respectively.

\paragraph{Structure-Guided Reconstruction.} We further use the generated sparse structure to condition reconstruction. Each voxel center \(x_j\) is encoded with Fourier positional encoding and an MLP to obtain a structure feature \(S_j\). Reconstruction features then query these structure features through cross-attention:

\begin{equation}
\label{eq:details-6}
S_j=\operatorname{MLP}(\operatorname{PE}(x_j)),
\qquad
R'=R+\operatorname{CA}(R,S,S).
\end{equation}

Here, \(S\) is the sequence of all structure features. We insert this operation into the VGGT backbone and point decoder so that point-map prediction uses both image observations and the complete structure. The network outputs point maps in the shared object space, confidence maps, and multi-level reconstruction features for subsequent generation.

\paragraph{Structure-Guided Shape Generation.}
Given a sparse structure and multi-view input images, we generate a detailed mesh through Structure-Guided Reconstruction and voxel-wise view conditioning. The sparse structure defines the spatial locations of the shape latent and also serves as a geometric condition for the reconstruction network. Conditioned on this structure, the reconstruction network predicts per-view point maps in the shared object space, confidence maps, and multi-level reconstruction features. These outputs provide both geometric correspondences for view selection and image-aligned features for conditioning shape generation.

For each input view $v$, we fuse the multi-level reconstruction features with DINOv3 image features to construct the generation condition:
\begin{equation}
    C_v = D_v + \psi\!\left(
        \operatorname{Concat}_{l}\!\left[F_v^{l}\right]
    \right),
    \label{eq:shape_feature_condition}
\end{equation}
where $F_v^{l}$ denotes the reconstruction features of view $v$ at feature level $l$, $D_v$ denotes the corresponding DINOv3 image features, and $\psi$ is a feature adapter that performs multi-level feature mapping, fusion, and spatial alignment. The resulting condition combines reconstruction features informed by the sparse structure with visual features extracted from the input image.

We use the predicted point maps to establish pixel-to-voxel
correspondences in the shared anchor-guided space. For each view,
we retain the highest-confidence 95\% of valid foreground
predictions and quantize the predicted points onto the voxel grid.

For a voxel centered at $x_j$, we group the retained pixels
assigned to it in each view $v$ and define the support score
$s_v(x_j)$ as the number of pixels in their largest connected
component. For voxels with direct support, we select
\begin{equation}
v_j^\star = \arg\max_v s_v(x_j).
\end{equation}
Ties favor the anchor view, followed by higher mean confidence
within the selected component.

A voxel without direct support inherits the selected view
of the most strongly supported voxel in its 26-neighborhood.
If the neighborhood also provides no support, we use the
anchor view, which is the first input view.

During shape generation, DiT alternates between multi-view attention and selected-view attention. Multi-view attention accesses the feature sequences of all input views, providing global context for shape generation. Selected-view attention allows each voxel to access the complete feature sequence $C_{v_j^{*}}$ of its assigned view. Thus, the selected-view condition is specific to each voxel while retaining the full visual context of the corresponding image. By alternating these two conditioning schemes, the model combines global multi-view information with voxel-associated visual evidence to recover geometric details. Under these conditions, DiT progressively updates the noisy shape latent at the sparse locations defined by the structure. The resulting latent is finally decoded into a mesh by the shape decoder.

\paragraph{Shape-Conditioned Texture Generation.} We encode the final mesh into a shape latent and concatenate it with the noisy texture latent at the same sparse locations as input to the texture DiT. The texture model follows the Structure-Guided Reconstruction and view-selection scheme of the shape stage, with an independent adapter for constructing image conditions. During generation, the shape latent remains fixed while the texture latent is progressively updated. The resulting latent is decoded into base color, metallic, roughness, and opacity attributes and baked into UV textures.

\subsection{More Results}
\label{app:results}

\subsubsection{Qualitative Ablation of Structure Conditioning}
\label{app:structure_conditioning}

\begin{figure}[!t]
    \centering

    \includegraphics[width=\columnwidth]
    {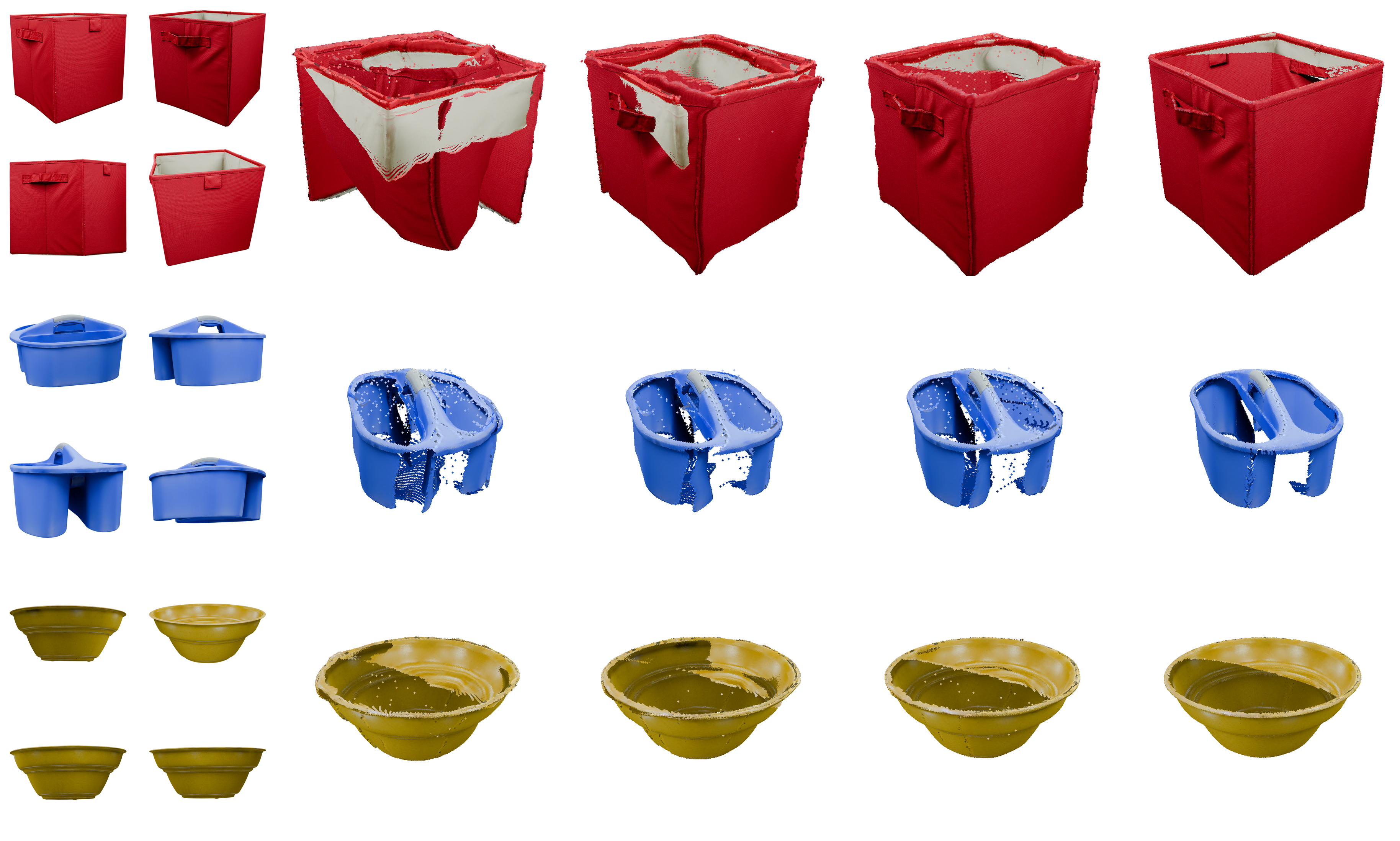}

    \vspace{-1mm}

    \noindent\makebox[\columnwidth][l]{%
        \parbox{0.20\columnwidth}{\centering\scriptsize Inputs}%
        \parbox{0.20\columnwidth}{\centering\scriptsize Image-only}%
        \parbox{0.20\columnwidth}{\centering\scriptsize Joint}%
        \parbox{0.20\columnwidth}{\centering\scriptsize Ours}%
        \parbox{0.20\columnwidth}{\centering\scriptsize GT}%
    }

\caption{
    \textbf{Qualitative comparison of point-map reconstruction.}
    Each row shows one object. From left to right: four input views,
    point maps predicted by image-only VGGT, jointly trained VGGT,
    and structure-conditioned VGGT (ours), followed by ground-truth
    visible point maps.
}
    \label{fig:vggt_structure_conditioning}
\end{figure}

Fig.~\ref{fig:vggt_structure_conditioning} presents additional qualitative
comparisons of different VGGT conditioning strategies. Each row corresponds
to one object. From left to right, we show four input views, point maps predicted
by image-only VGGT, jointly trained VGGT, and structure-conditioned
VGGT (ours), followed by the ground-truth visible point maps. Joint training improves geometric consistency over the image-only baseline.
Furthermore, explicitly conditioning VGGT on the generated sparse 3D structure
produces more complete and coherent geometry.

\subsubsection{Qualitative Ablation of View Selection and Conditioning}
\label{app:view_conditioning}
\begin{figure}[!t]
    \centering
    \includegraphics[width=\linewidth]{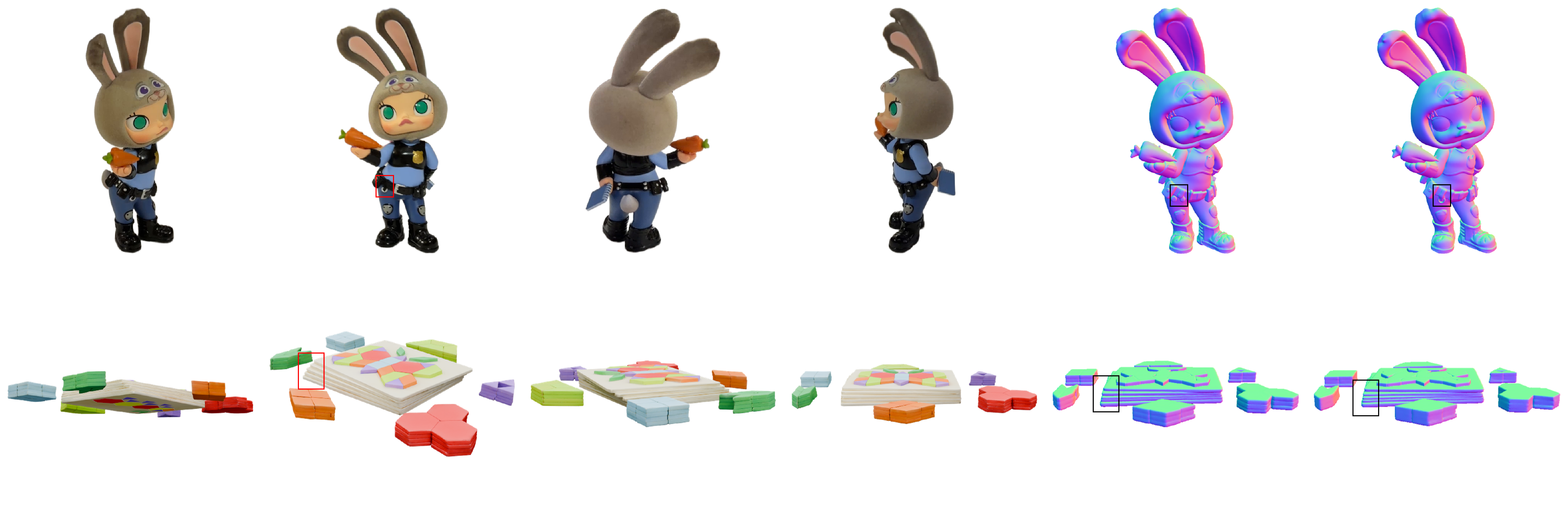}

    \smallskip
    \noindent\makebox[\linewidth][l]{%
        \parbox{0.666\linewidth}{\centering\scriptsize Input Views}%
        \parbox{0.167\linewidth}{\centering\scriptsize
            Full Multi-View\\Conditioning}%
        \parbox{0.167\linewidth}{\centering\scriptsize Ours}%
    }

    \caption{
        \textbf{Qualitative ablation of voxel-wise view selection
        and conditioning.}
        Each row shows four input views followed by rendered normal
        maps of shapes generated with full multi-view conditioning
        and our method, respectively.
        Both variants use the same predicted sparse structure.
        Red boxes highlight reference details in the input images,
        and black boxes mark the corresponding regions in the
        generated shapes.
    }
    \label{fig:view_conditioning_qualitative}
\end{figure}

Fig.~\ref{fig:view_conditioning_qualitative} provides qualitative
results complementing Table~\ref{tab:view_conditioning_ablation_gso}.
Using the same predicted sparse structure, we compare full
multi-view conditioning with our alternating full multi-view and
selected-view conditioning. The highlighted regions show that our method better preserves local geometric details, demonstrating the benefit of introducing view-specific visual evidence.

\subsubsection{Additional Qualitative Results on Rendered Assets}
\label{app:rendered_assets}

\begin{figure*}[t]
    \centering
    \begin{minipage}{0.92\textwidth}
        \centering
        \includegraphics[width=\linewidth]{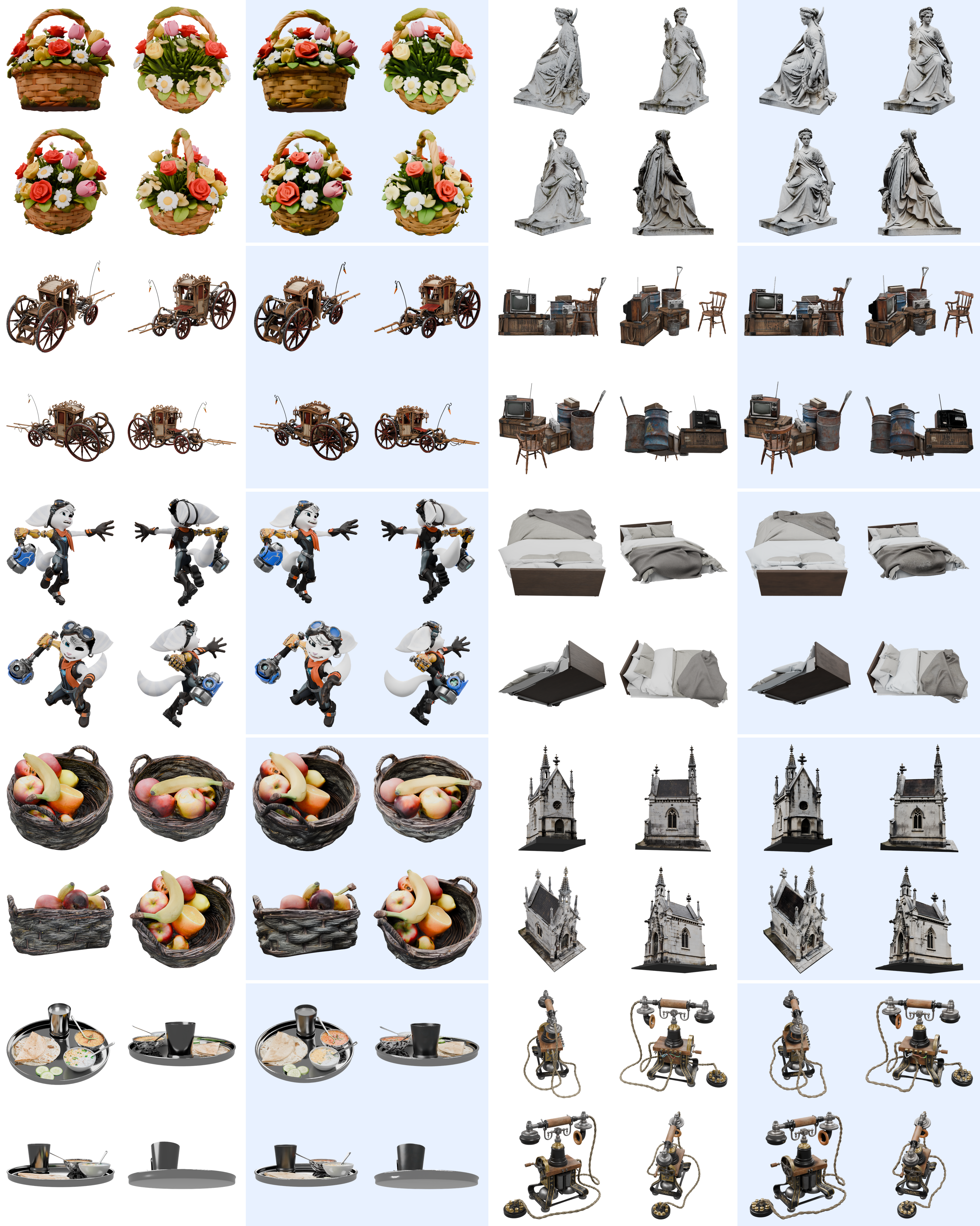}

        \noindent
        \makebox[\linewidth][c]{%
            \parbox{0.25\linewidth}{\centering\textbf{Input}}%
            \parbox{0.25\linewidth}{\centering\textbf{Ours}}%
            \parbox{0.25\linewidth}{\centering\textbf{Input}}%
            \parbox{0.25\linewidth}{\centering\textbf{Ours}}%
        }
    \end{minipage}

    \caption{
        \textbf{Additional qualitative results on rendered 3D assets.}
        Each group shows four input views on a white background
        and the corresponding renderings of our generated mesh
        on a light-blue background.
        For direct visual comparison, we geometrically align
        the generated mesh with the ground-truth asset and
        render it using the same cameras as the input views.
    }
    \label{fig:qualitative_rendered_assets}
\end{figure*}

Fig.~\ref{fig:qualitative_rendered_assets} presents additional
results on rendered 3D assets. For comparison with the input
observations, we geometrically align each generated mesh with
the ground-truth asset and render it from the input cameras.
These examples illustrate how our method preserves the
observed geometry and appearance across different objects.

\subsubsection{Visualization of Voxel-Wise View Selection}
\label{app:voxel_view_selection}

\begin{figure}[!t]
    \centering
    \includegraphics[width=\linewidth]{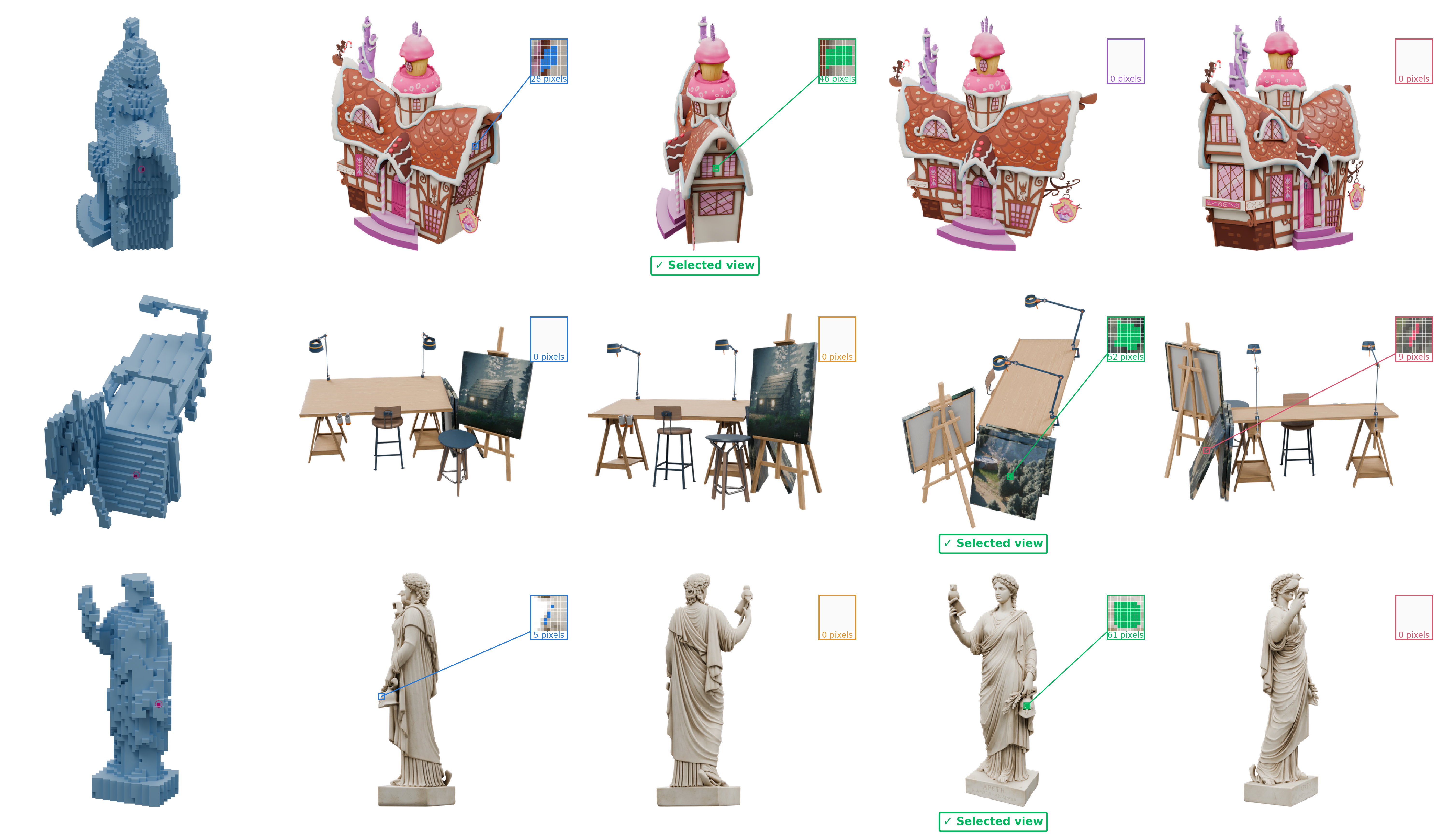}
    \caption{\textbf{Visualization of voxel-wise view selection
    from our inference results.}
    Each row shows a generated sparse structure with a query voxel
    highlighted in magenta, followed by four input views.
    Colored pixels and enlarged insets show the predicted
    pixel-to-voxel correspondences, with numbers indicating
    highlighted pixels within each inset.
    Green checkmarks identify the views selected by our model.}
    \label{fig:voxel_view_selection}
\end{figure}

Fig.~\ref{fig:voxel_view_selection} visualizes voxel-wise view selection
from our model's actual inference results, including the generated sparse
structures, predicted pixel-to-voxel correspondences, and selected views.
Pixel support measures the amount of image evidence associated with a
local 3D region and serves as a practical proxy for view informativeness.
In these examples, the selected views expose the queried surface regions
more directly and provide stronger pixel support than the alternatives.
These results qualitatively support our use of pixel-to-voxel
correspondences to identify informative views and provide region-specific
visual evidence for detailed shape generation.

\subsubsection{Shared-Space Consistency between Reconstruction and Generation}
\label{sec:shared_space_consistency}

\begin{figure*}[t]
    \centering
    \includegraphics[width=\textwidth]
    {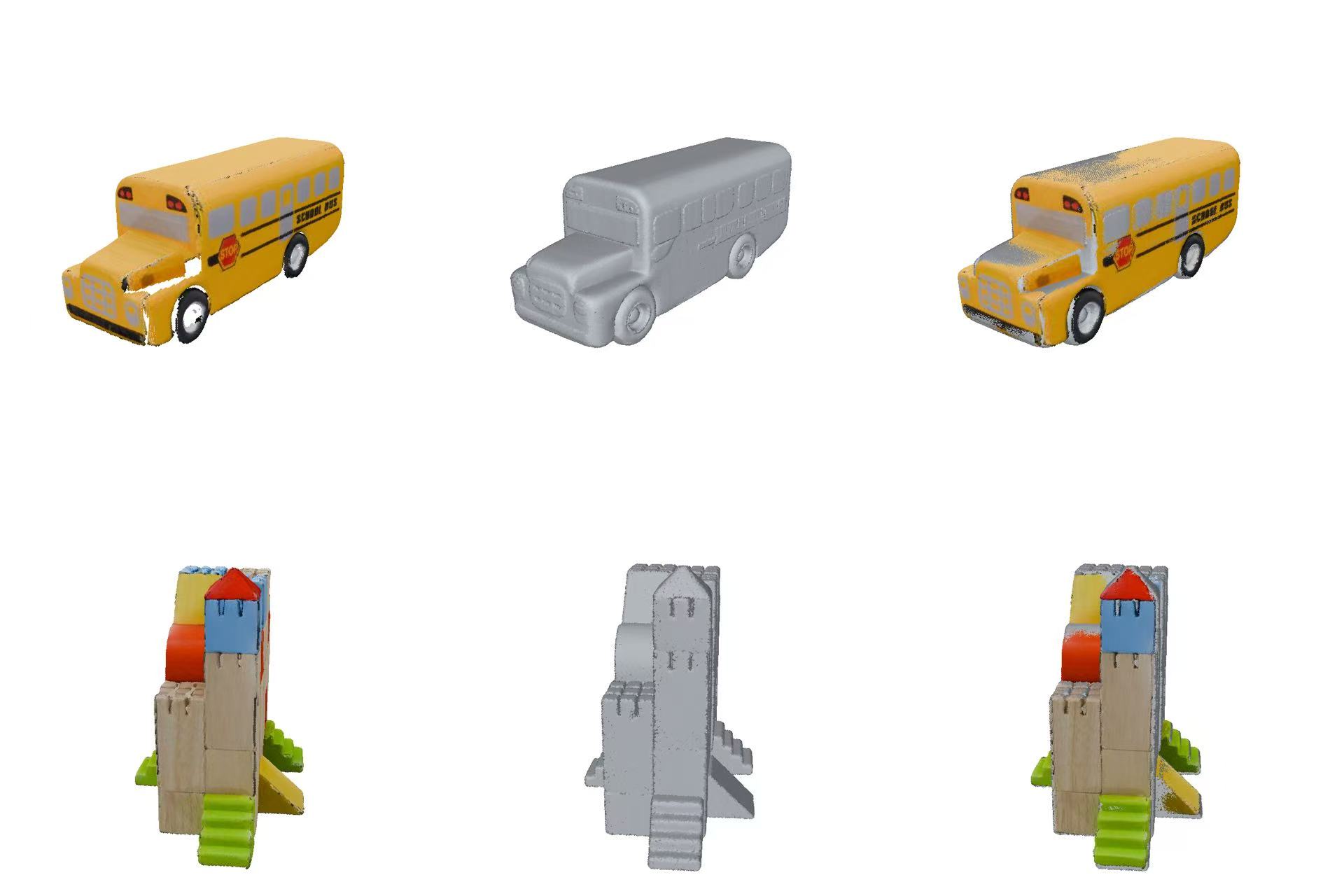}
    \caption{
    Shared-space consistency of \textit{Structure-Conditioned VGGT}.
    Each row shows the reconstructed colored point cloud, the generated
    mesh, and their direct overlay from left to right.
    No object-specific alignment or registration is applied.
    }
    \label{fig:ss_conditioned_point_mesh}
\end{figure*}

\begin{figure*}[t]
    \centering
    \includegraphics[width=\textwidth]
    {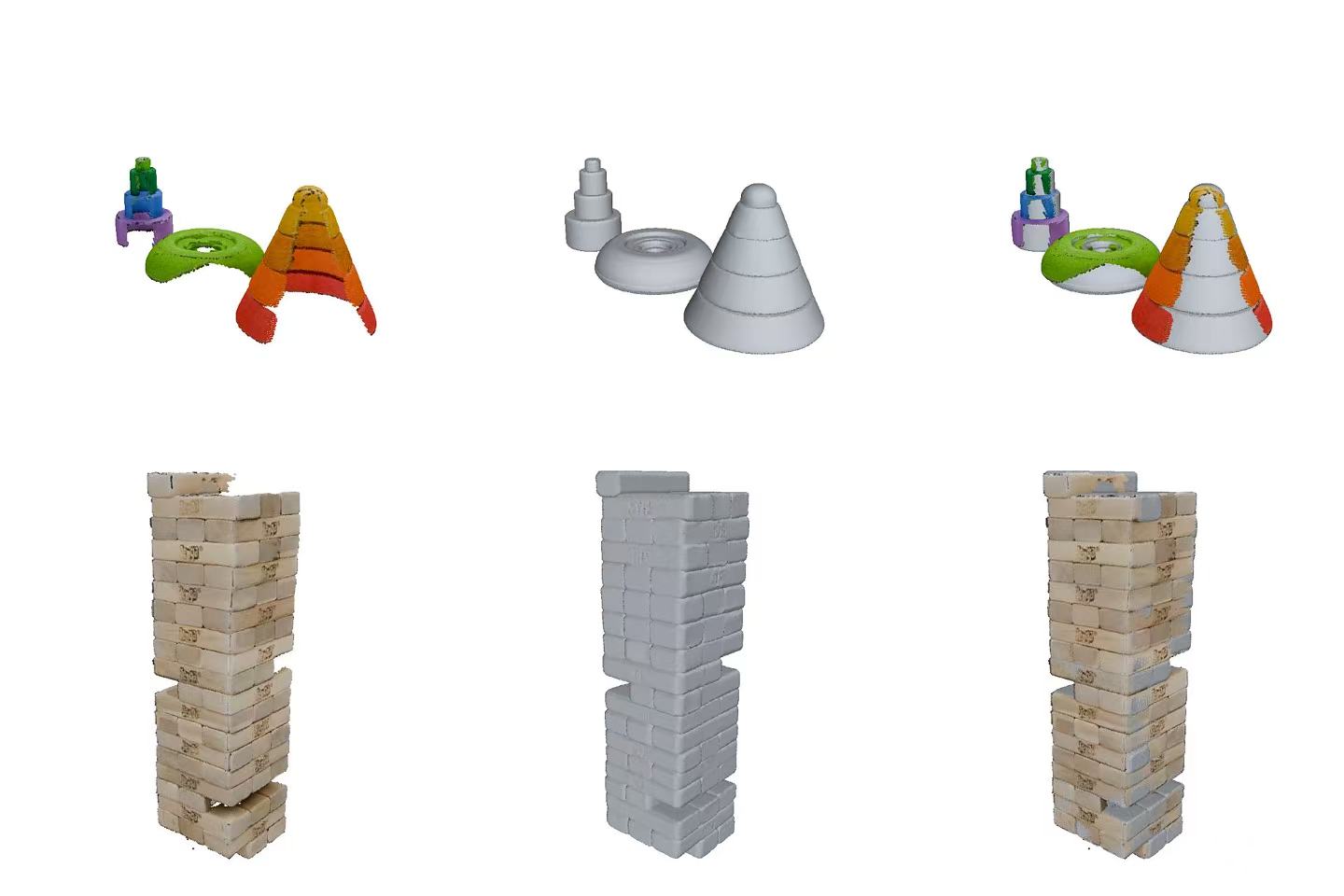}
    \caption{
    Shared-space consistency of \textit{Jointly Trained VGGT}.
    Each row shows the reconstructed colored point cloud, the generated
    mesh, and their direct overlay from left to right.
    No object-specific alignment or registration is applied.
    }
    \label{fig:joint_point_mesh}
\end{figure*}

Figures~\ref{fig:ss_conditioned_point_mesh} and
\ref{fig:joint_point_mesh} visualize the reconstructed point clouds
together with the generated meshes for \textit{Structure-Conditioned
VGGT} and \textit{Jointly Trained VGGT}, respectively.
All results are obtained directly from inference and shown in the shared
model coordinate system, without object-specific alignment or
registration. The close correspondence in the overlays demonstrates
that the reconstructed point clouds and generated meshes reside in the
same space and are directly spatially aligned.

\end{document}